# RMNA: A Neighbor Aggregation-Based Knowledge Graph Representation Learning Model Using Rule Mining

Ling Chen[1]*  Jun Cui[1]  Xing Tang[1]  Chaodu Song[1]  Yuntao Qian[1]  Yansheng Li[2]
Yongjun Zhang[2]

[1]*College of Computer Science and Technology, Zhejiang University, Hangzhou 310027, China*
[2]*School of Remote Sensing and Information Engineering, Wuhan University, Wuhan 430079, China*

**Abstract**: Although the state-of-the-art traditional representation learning (TRL) models show competitive performance on knowledge graph completion, there is no parameter sharing between the embeddings of entities, and the connections between entities are weak. Therefore, neighbor aggregation-based representation learning (NARL) models are proposed, which encode the information in the neighbors of an entity into its embeddings. However, existing NARL models either only utilize one-hop neighbors, ignoring the information in multi-hop neighbors, or utilize multi-hop neighbors by hierarchical neighbor aggregation, destroying the completeness of multi-hop neighbors. In this paper, we propose a NARL model named RMNA, which obtains and filters horn rules through a rule mining algorithm, and uses selected horn rules to transform valuable multi-hop neighbors into one-hop neighbors, therefore, the information in valuable multi-hop neighbors can be completely utilized by aggregating these one-hop neighbors. In experiments, we compare RMNA with the state-of-the-art TRL models and NARL models. The results show that RMNA has a competitive performance.

---

* Corresponding author. Tel: +86 13606527774.
*E-mail address: lingchen@cs.zju.edu.cn (Ling Chen), cuijun@cs.zju.edu.cn (Jun Cui), tangxing@cs.zju.edu.cn (Xing Tang), songcd2020@zju.edu.cn (Chaodu Song), ytqian@zju.edu.cn (Yuntao Qian), yansheng.li@whu.edu.cn (Yansheng Li), zhangyj@whu.edu.cn (Yongjun Zhang).*

# 1 INTRODUCTION

Knowledge graphs (KGs) are directed graphs where nodes represent entities, and edges represent relations, and have been applied to many NLP tasks, e.g., relation extraction [14], question answering [31], information retrieval [10], semantic similarity measure [26], and recommendation [9]. Figure 1 shows a fragment of a KG. KGs contain structural human knowledge, each of which can be represented as a triple (head entity, relation, tail entity) or $(h, r, t)$, indicating the relationship between two entities. For example, (William Shakespeare, masterpiece, Hamlet) shows that the "masterpiece" of "William Shakespeare" is "Hamlet". Although existing large-scale knowledge graphs, e.g., Freebase [3], DBPedia [2], Yago [20], and WordNet [16], have included a large amount of knowledge, they are still far from complete. Therefore, there are a lot of researches focusing on knowledge graph completion (KGC), which aims at completing missing information in KGs. Link prediction is a subtask of KGC, which predicts the corresponding tail (head) entity given a head (tail) entity and a relation.

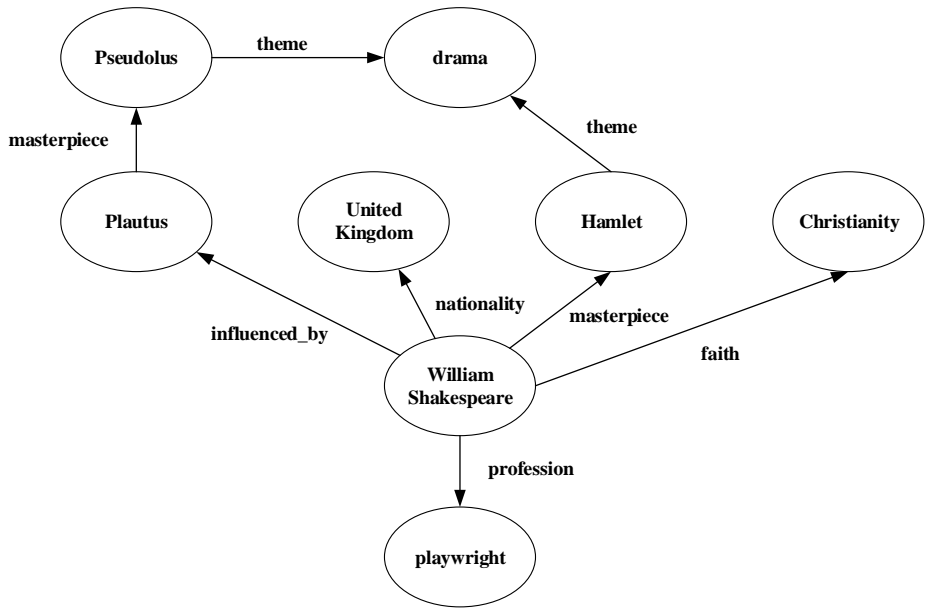

Figure 1: A fragment of a knowledge graph.

Recently, representation learning, which embeds relations and entities into a low-dimensional vector space [6], has been the most popular way to represent relations and entities in KGs, and the state-of-the-art models of KGC are mostly based on representation learning. These models can be divided into two categories: traditional representation learning (TRL) models, and neighbor aggregation-based representation learning (NARL) models.

TRL models are composed of operations that do not support parameter sharing between the embeddings of entities, therefore, there is little information interaction between entities, and the connections between entities are weak. For example, TransE [4] regards the relation as a translation between the entities in a given triple, which is composed of addition and subtraction operations; DistMult [29] aims to minimize the inner product of the embeddings of the entities and the relation in a given triple.

The neighbor of an entity is a path-entity pair. The entity in a pair is called neighbor entity, which is reachable by one or more hops from the given entity. The path in a pair is called neighbor path (or neighbor relation if it consists of only one relation), which is the reachable path between the given entity and the neighbor entity. For example, in Figure 1, ((masterpiece), Hamlet) and ((masterpiece, theme), drama) are both neighbors of "William Shakespeare", where the former is one-hop reachable from "William Shakespeare", i.e., a one-hop neighbor, and the latter is multi-hop (two-hop) reachable from "William Shakespeare", i.e., a multi-hop neighbor.

Neighbor aggregation aims at aggregating the neighbors of an entity, and forming the neighbor-based embedding of the entity. Therefore, the embeddings of entities in NARL models share parameters, and the connections between entities are strong. For example, both A2N [1] and LAN [27] introduce a neighbor aggregation model to utilize one-hop neighbors. However, they ignore the information in multi-hop neighbors. Since the number of neighbors increases exponentially with the number of hops, directly aggregating all multi-hop neighbors of an entity causes a large amount of calculation. Therefore, R-GCN [19] and KBGAT [17] introduce graph neural networks (GNNs) to hierarchically aggregate neighbors, i.e., each layer only aggregates one-hop neighbors, and multiple layers are introduced to utilize multi-hop neighbors. However, this approach splits a complete multi-hop neighbor into several one-hop neighbors, destroying the completeness of multi-hop neighbors.

To address the aforementioned problems, we propose a novel NARL model named RMNA. RMNA obtains horn rules by rule mining algorithm AMIE, filters out selected horn rules, and uses them to transform valuable multi-hop neighbors into one-hop neighbors with similar semantics. Then, RMNA separately and hierarchically aggregates the original one-hop neighbors and the transformed one-hop neighbors of an entity by the multi-head attention mechanism [24], where the embeddings of both original one-hop neighbors and transformed one-hop neighbors are considered, and some factors that can measure the reliabilities of the corresponding selected horn rules are considered for transformed one-hop neighbors. Then, the heads are aggregated by the self-attention mechanism [24] to form neighbor-based embeddings. In this way, the information in valuable multi-hop neighbors can be completely utilized by aggregating transformed one-hop neighbors. The main contributions of RMNA are summarized as follows:

● Propose RMNA, which transforms valuable multi-hop neighbors into one-hop neighbors that are semantically similar to the corresponding multi-hop neighbors, so that the completeness of multi-hop neighbors can be ensured.

● Introduce a hierarchical neighbor aggregation model, which separately aggregates the original one-hop neighbors and the transformed one-hop neighbors of an entity, so that the information in two types of one-hop neighbors can be learned effectively.

● Compare RMNA with the state-of-the-art TRL models and NARL models on two datasets. Experiment results show that RMNA has a competitive performance.

## 2    RELATED WORK

In this section, we introduce two lines of works related to our model: TRL models and NARL models.

### 2.1    TRL Models

In TRL models, each relation or entity usually has only one embedding, which is learned by directly applying an energy function on triples in KGs. The energy functions in TRL models are composed of operations that do not support parameter sharing between entities.

Translation models are typical TRL models. TransE [4] regards the relation as a translation between the entities in a given triple, i.e., the difference between the embeddings of the tail entity and the head entity should be close to the embedding of the relation. With simple operations and low calculation, TransE shows a competitive performance. Based on TransE, a lot of improvement methods are proposed [7; 11; 12; 15; 21; 28; 31]. For example, TransH [28] learns a hyperplane for each relation, and projects the entities into the relation-specific hyperplane in a given triple. Lin et al. [15] believe relations and entities belonging to different semantic spaces, and proposed TransR, which projects the entities into the relation-specific space in a given triple. TransD [12] learns two embeddings for each relation and entity to construct mapping matrices, reducing calculation compared to TransR. Fan et al. [7] proposed a probabilistic model IKE to measure the probability of each belief.

In addition to translation-based models, there are other TRL models. DistMult [29] is a bilinear model, which aims to minimize the inner product of the embeddings of the entities and the relation in a given triple. Since some relations in KGs are asymmetric, and the existing models cannot properly handle the asymmetry of the relations, ComplEx [22] learns a complex embedding for each relation and entity. To increase the expressiveness of embeddings, ConvE [5] transforms the embeddings of the head entity and the relation in a triple into 2D matrices and concatenates them into an input matrix, which is then applied to a convolutional layer. Nguyen et al. [18] point out that ConvE ignores local relationships among different dimension entries, and proposed ConvKB, which uses multiple convolution kernels to obtain features among different dimension entries.

However, since there is no parameter sharing between entities in TRL models, there is little information interaction between entities, and the connections between entities are weak.

### 2.2 NARL Models

In NARL models, each relation and entity has a base embedding and a neighbor-based embedding, which is obtained by neighbor aggregation on the base embeddings. Therefore, the embeddings of entities in NARL models share parameters, and the connections between entities are strong. NARL models consist of an encoder, i.e., a neighbor aggregation model, which aggregates the neighbors of an entity to form its neighbor-based embedding, and a decoder, which trains a TRL model using the neighbor-based embeddings as inputs. Existing NARL models can be divided into models only utilizing one-hop neighbors and models utilizing multi-hop neighbors.

A2N [1] and LAN [27] are typical models only utilizing one-hop neighbors, and are both query-based models, i.e., learning a different neighbor-based embedding for each entity under each different relation, which is called query relation. Given an entity and a query relation, A2N aggregates the one-hop neighbors of the given entity by a bilinear attention, which is conditioned on the query relation; LAN proposes a neighbor aggregation model that considers data mining features, which uses the co-occurrences of the query relation and the neighbor relation as a part of the attention mechanism. However, both A2N and LAN ignore the information in multi-hop neighbors.

Since the number of neighbors increases exponentially with the number of hops, directly aggregating all multi-hop neighbors of an entity causes a large amount of calculation. Therefore, models utilizing multi-hop neighbors aggregate neighbors by multi-layer GNNs. R-GCN [19] classifies the neighbors of an entity according to the type and direction of the neighbor relations, and uses graph convolutional network (GCN) as its encoder. Since the neighbors in the same classification have the same attention in R-GCN, Nathani et al. [17] proposed KBGAT, which uses Graph Attention Network (GAT) [25] as its encoder, and calculates the attentions of each neighbor. However, both R-GCN and KBGAT split a complete multi-hop neighbor into several one-hop neighbors, destroying the completeness of multi-hop neighbors.

## 3 METHODOLOGY

In this section, we introduce the details of RMNA, which consists of three parts: the rule mining part, the neighbor aggregation part, and the representation learning part.

### 3.1 Preliminaries

We use $\mathcal{E}$ and $\mathcal{R}$ to represent entities and relations in the KG, respectively. The KG consists of triples $(h, r, t)$, where $h, t \in \mathcal{E}$ and $r \in \mathcal{R}$, therefore, it can be represented as $\text{KG} = \{(h, r, t)\}$. The target of RMNA is link prediction, i.e., completing a triple whose $h$ or $t$ is missing.

RMNA obtains and filters horn rules, and uses selected horn rules to transform valuable multi-hop neighbors into one-hop neighbors. The definition of the horn rule is given as follows.

**Definition 1 (atom):** An atom is a fact represented in the form of $r(e, e')$, which can be seen as another representation of a triple $(e, r, e')$. Both $e$ and $e'$ are entity variables.

**Definition 2 (horn rule):** A horn rule is a rule in the form of $B_1 \wedge B_2 \ldots \wedge B_n \to r(e, e')$, where $\{B_1, B_2, \ldots, B_n\}$ is a set of atoms named the body, and $r(e, e')$ is an atom named the head, which can be abbreviated as $B \to r(e, e')$. In this paper, we only consider closed horn rules where all entity variables appear at least twice, e.g., $FathorOf(e, e_1) \wedge MotherOf(e', e_1) \to MarriedTo(e, e')$.

### 3.2 Rule Mining Part

In the rule mining part, RMNA completes the following three tasks: base embedding initialization, rule mining and filtering, rule matching.

#### 3.2.1 Base Embedding Initialization

RMNA initializes the base embeddings of relations and entities on the original KG by TransE [4]. Given a triple $(h, r, t)$, TransE regards the relation $r$ as a translation between the head entity $h$ and the tail entity $t$, i.e., $\mathbf{h} + \mathbf{r} \approx \mathbf{t}$. The energy function of TransE is defined as follows:

$$E_{\text{TransE}}(h, r, t) = |\mathbf{h} + \mathbf{r} - \mathbf{t}|. \tag{1}$$

#### 3.2.2 Rule Mining and Filtering

RMNA obtains horn rules by rule mining algorithm AMIE [8], and filters out selected horn rules that can transform valuable multi-hop neighbors into one-hop neighbors with similar semantics. We set the maximum length of the horn rules to be $l_{\max}$, i.e., the maximum number of atoms in the bodies of horn rules is $l_{\max}$.

AMIE is a classical rule mining algorithm on KGs, which can search horn rules and provide metrics such as confidence and head coverage. The support of a horn rule $B \to r(e, e')$ represents the number of times that the body $B$ and the head $r(e, e')$ are satisfied simultaneously in the KG, which is defined as follows:

$$sup(B \to r(e, e')) = |\{(e_1, e_2) | \exists z_1, \ldots, z_m : B \wedge r(e_1, e_2), e_1, e_2, z_1, \ldots, z_m \in \mathcal{E}\}|, \tag{2}$$

where $|x|$ is the count of $x$; $z_1, \ldots, z_m$ are entity variables of $B$ except $e_1$ and $e_2$. The head coverage of a horn rule $B \to r(e, e')$ is defined as follows:

$$hc(B \rightarrow r(e,e')) = \frac{sup(B \rightarrow r(e,e'))}{|\{(e_1,e_2)|r(e_1,e_2), e_1, e_2 \in \mathcal{E}\}|}, \quad (3)$$

which represents the probability that the body $B$ and the head $r(e,e')$ are satisfied simultaneously when the head $r(e,e')$ is already satisfied. The confidence of a horn rule $B \rightarrow r(e,e')$ is defined as follows:

$$conf(B \rightarrow r(e,e')) = \frac{sup(B \rightarrow r(e,e'))}{|\{(e_1,e_2)|\exists z_1,\ldots,z_m: B, z_1,\ldots,z_m \in \mathcal{E}\}|}, \quad (4)$$

which represents the probability that the body $B$ and the head $r(e,e')$ are satisfied simultaneously when the body $B$ is already satisfied. Both head coverage and confidence can measure the reliability of a horn rule.

After rule mining on KGs by the AMIE algorithm, RMNA filters horn rules through the following two steps: (1) filtering out the horn rules in the form of $r_1(e,e_1) \wedge r_2(e_1,e_2) \wedge \ldots \wedge r_n(e_{n-1},e') \rightarrow r(e,e')$ whose body is a path from $e$ to $e'$, and head is an atom with $e$ and $e'$, as they can transform multi-hop neighbors into one-hop neighbors; (2) filtering out horn rules whose head coverages and confidences are greater than thresholds $hc_{min}$ and $conf_{min}$, respectively, to control the quantity and quality of the horn rules.

### 3.2.3 Rule Matching

RMNA transforms valuable multi-hop neighbors into one-hop neighbors by selected horn rules. Specifically, RMNA matches each entity with each selected horn rule: if an entity contains a multi-hop neighbor that has the same form with the body of a selected horn rule, then introducing the one-hop neighbor that has the same form with the head of the horn rule for the entity.

For example, given a selected horn rule " $masterpiece(e,e_1) \wedge theme(e_1,e') \rightarrow creative\_style(e,e')$ " and an entity "William Shakespeare" in Figure 1, "William Shakespeare" contains a multi-hop neighbor ((masterpiece, theme), drama), which has the same form with the body of the given selected horn rule, therefore, RMNA introduces the one-hop neighbor ((creative_style), drama) for "William Shakespeare", which has the same form with the head of the given selected horn rule.

## 3.3 Neighbor Aggregation Part

In the neighbor aggregation part, RMNA learns neighbor-based embeddings for each entity in the KG through a hierarchical neighbor aggregation model. Figure 2 shows the structure of the hierarchical neighbor aggregation model.

### 3.3.1 Overall Architecture

Given an entity $e$, its one-hop neighbors can be divided into original one-hop neighbors, which are existed in the original KG, and transformed one-hop neighbors, which are transformed from valuable multi-hop neighbors. RMNA separately aggregates original one-hop neighbors and transformed one-hop neighbors, so that the information in two types of one-hop neighbors can be learned effectively.

In each layer, firstly, we construct the inputs for two types of one-hop neighbors. Then, the multi-head attention mechanism with $K_m$ heads calculates the weight of each original one-hop neighbor and transformed one-hop neighbor, and aggregates them separately, forming hidden vectors. At last, the self-attention mechanism aggregates the hidden vectors, forming the neighbor-based embedding of the given entity $e$ in this layer.

After two layers of neighbor aggregation, finally, we obtain the neighbor-based embedding of the given entity $e$. We use the energy function of TransE as the training target on neighbor-based embeddings to update the parameters in the neighbor aggregation part.

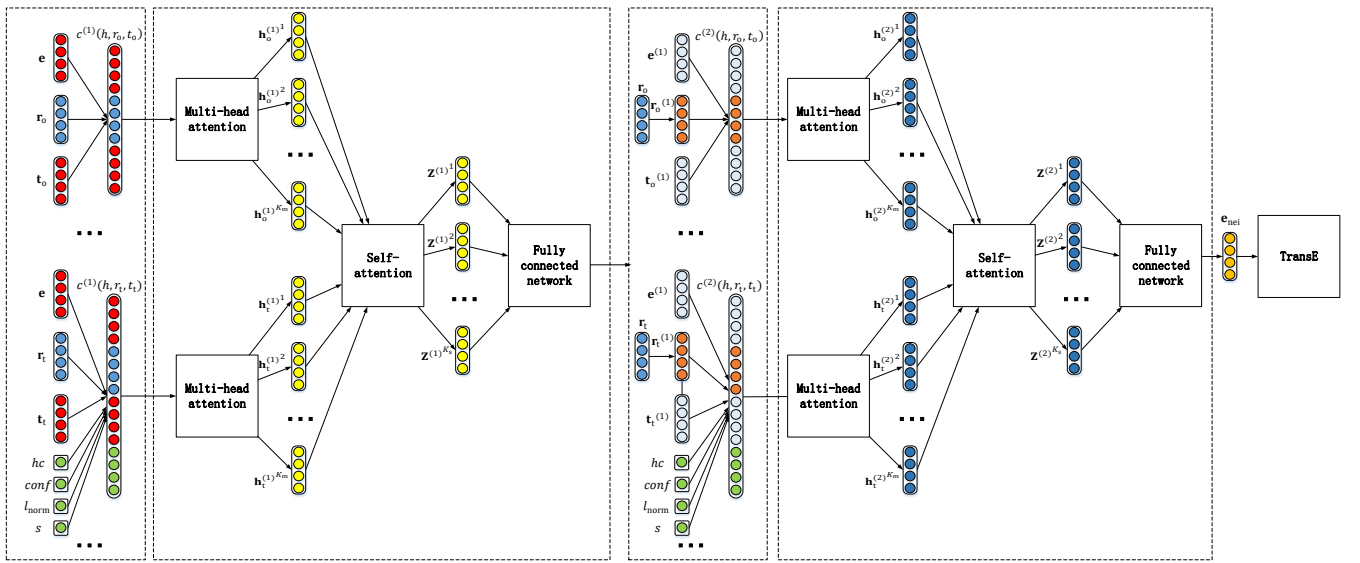

Figure 2: The structure of the hierarchical neighbor aggregation model.

### 3.3.2 Input Construction

We design different forms of inputs for two types of one-hop neighbors: since the correctness of original one-hop neighbors can be guaranteed, we only consider the embeddings of the given entity, neighbor relation, and neighbor entity; while transformed one-hop neighbors are transformed from valuable multi-hop neighbors, and their correctness cannot be guaranteed, we consider additional factors to measure their reliabilities.

In the first layer, we construct the inputs for two types of one-hop neighbors associated with the given entity $e$ through the following formulas:

$$c^{(1)}(h, r_o, t_o) = \mathbf{W}_{c,o}^{(1)}[\mathbf{e}, \mathbf{r}_o, \mathbf{t}_o], \tag{5}$$

$$c^{(1)}(h, r_t, t_t) = \mathbf{W}_{c,t}^{(1)}[\mathbf{e}, \mathbf{r}_t, \mathbf{t}_t, hc, conf, l_{norm}, s], \tag{6}$$

where $\mathbf{W}_{c,o}^{(1)} \in \mathbb{R}^{d^{(1)} \times 3d}$ and $\mathbf{W}_{c,t}^{(1)} \in \mathbb{R}^{d^{(1)} \times (3d+4)}$ are the linear transformation matrices of original one-hop neighbors and transformed one-hop neighbors in the first layer, respectively; $[, ..., ]$ is the concatenation of the embeddings; $\mathbf{e}, \mathbf{r}_o, \mathbf{t}_o, \mathbf{r}_t, \mathbf{t}_t \in \mathbb{R}^d$ are the base embeddings of $e$, $r_o$, $t_o$, $r_t$ and $t_t$, respectively; $hc$ and $conf$ are the head coverage and confidence of the corresponding horn rule, respectively; $l_{norm}$ is the normalized length of the corresponding horn rule, i.e., the ratio of the length of the horn rule to $l_{max}$; $s$ is the score of $(r_t, t_t)$ calculated by the energy function of TransE, which can measure the fit of the transformed one-hop neighbor to the original KG and is defined as follows:

$$s(e, r_t, t_t) = |\mathbf{e} + \mathbf{r}_t - \mathbf{t}_t|. \tag{7}$$

In the second layer, the inputs for two types of one-hop neighbors are defined as follows:

$$c^{(2)}(h, r_o, t_o) = \mathbf{W}_{c,o}^{(2)}[\mathbf{e}^{(1)}, \mathbf{r}_o^{(1)}, \mathbf{t}_o^{(1)}], \tag{8}$$

$$c^{(2)}(h, r_t, t_t) = \mathbf{W}_{c,t}^{(2)}[\mathbf{e}^{(1)}, \mathbf{r}_t^{(1)}, \mathbf{t}_t^{(1)}, hc, conf, l_{norm}, s], \tag{9}$$

where $\mathbf{W}_{c,o}^{(2)} \in \mathbb{R}^{d^{(2)} \times 3d^{(1)}}$ and $\mathbf{W}_{c,t}^{(2)} \in \mathbb{R}^{d^{(2)} \times (3d^{(1)}+4)}$ are the linear transformation matrices of original one-hop neighbors and transformed one-hop neighbors in the second layer, respectively; $\mathbf{e}^{(1)}, \mathbf{t}_o^{(1)}, \mathbf{t}_t^{(1)} \in \mathbb{R}^{d_1}$ are the neighbor-based embeddings of $e$, $t_o$, and $t_t$ in the first layer, respectively; $\mathbf{r}_o^{(1)} = \mathbf{r}_o \mathbf{W}_r$, $\mathbf{r}_t^{(1)} = \mathbf{r}_t \mathbf{W}_r$ are the transformed embeddings of $r_o$ and $r_t$, respectively; $\mathbf{W}_r \in \mathbb{R}^{d^{(1)} \times d}$ is the relation transformation matrix.

### 3.3.3 Multi-Head Attention Mechanism

Receiving the inputs of two types of one-hop neighbors, the multi-head attention mechanism, which consists of two multi-head attention models, then measures the weight of each one-hop neighbor. The attentions of each original one-hop neighbor and transformed one-hop neighbor on the $k$-th head in the $l$-th layer are defined as follows:

$$b^{(l)^k}(h, r_o, t_o) = LeakyReLU\left(\mathbf{W}_{b,o}^{(l)^k} c^{(l)}(h, r_o, t_o)\right), \tag{10}$$

$$b^{(l)^k}(h, r_t, t_t) = LeakyReLU\left(\mathbf{W}_{b,t}^{(l)^k} c^{(l)}(h, r_t, t_t)\right), \tag{11}$$

where $\mathbf{W}_{b,o}^{(l)^k}, \mathbf{W}_{b,t}^{(l)^k} \in \mathbb{R}^{1 \times d^{(l)}}$ are the weight matrices of original one-hop neighbors and transformed one-hop neighbors on the $k$-th head in the $l$-th layer, respectively.

We apply a $softmax$ function on the attentions of two types of one-hop neighbors separately to obtain normalized attention values, which are defined as follows:

$$\alpha^{(l)^k}(h, r_o, t_o) = \frac{\exp(b^{(l)^k}(h, r_o, t_o))}{\sum_{(r_o', t_o') \in N_o(e)} \exp(b^{(l)^k}(h, r_o', t_o'))}, \tag{12}$$

$$\alpha^{(l)^k}(h, r_t, t_t) = \frac{\exp(b^{(l)^k}(h, r_t, t_t))}{\sum_{(r_t', t_t') \in N_t(e)} \exp(b^{(l)^k}(h, r_t', t_t'))}, \tag{13}$$

where $N_o(e)$ is the set of original one-hop neighbors of $e$; $N_t(e)$ is the set of transformed one-hop neighbors of $e$

At last, two multi-head attention models aggregate the two types of one-hop neighbors separately, each forming $K_m$ hidden vectors, corresponding to $K_m$ heads. The hidden vectors of the two multi-head attention models on the $k$-th head in the $l$-th layer are defined as follows:

$$\mathbf{h}_o^{(l)^k} = \delta\left(\sum_{(r_o, t_o) \in N_o(e)} \alpha^{(l)^k}(h, r_o, t_o) \, c^{(l)}(h, r_o, t_o)\right) \in \mathbb{R}^{d^{(l)}}, \tag{14}$$

$$\mathbf{h}_t^{(l)^k} = \delta\left(\sum_{(r_t, t_t) \in N_t(e)} \alpha^{(l)^k}(h, r_t, t_t) \, c^{(l)}(h, r_t, t_t)\right) \in \mathbb{R}^{d^{(l)}}, \tag{15}$$

where $\delta$ is a non-linear function, e.g., ELU function.

### 3.3.4 Self-Attention Mechanism

The self-attention mechanism calculates the attentions of the $2K_m$ hidden vectors through a self-attention model with $K_s$ heads, and aggregates the hidden vectors, forming the neighbor-based embedding of the given entity $e$ through a fully connected neural network.

We concatenate the $2K_m$ hidden vectors $\{\mathbf{h}_o^{(l)^1}, \ldots, \mathbf{h}_o^{(l)^{K_m}}, \mathbf{h}_t^{(l)^1}, \ldots, \mathbf{h}_t^{(l)^{K_m}}\}$ into a matrix $\mathbf{X}^{(l)} \in \mathbb{R}^{2K_m \times d^{(l)}}$ as the input of the self-attention model. The query matrix, key matrix, and value matrix of the self-attention model on the $k$-th head in the $l$-th layer are defined as follows:

$$\mathbf{Q}^{(l)^k} = \mathbf{X}^{(l)} \mathbf{W}_Q^{(l)^k}, \tag{16}$$

$$\mathbf{K}^{(l)^k} = \mathbf{X}^{(l)} \mathbf{W}_K^{(l)^k}, \tag{17}$$

$$\mathbf{V}^{(l)^k} = \mathbf{X}^{(l)} \mathbf{W}_V^{(l)^k}, \tag{18}$$

where $\mathbf{W}_Q^{(l)^k} \in \mathbb{R}^{d^{(l)} \times d_Q^{(l)}}$, $\mathbf{W}_K^{(l)^k} \in \mathbb{R}^{d^{(l)} \times d_K^{(l)}}$, and $\mathbf{W}_V^{(l)^k} \in \mathbb{R}^{d^{(l)} \times d_V^{(l)}}$ ($d_Q^{(l)} = d_K^{(l)}$) are parameter matrices. Then, the output of the self-attention model on the $k$-th head in the $l$-th layer is defined as follows:

$$\mathbf{Z}^{(l)^k} = \text{softmax}\left(\frac{\mathbf{Q}^{(l)^k} \mathbf{K}^{(l)^{k^T}}}{\sqrt{d_K^{(l)}}}\right) \mathbf{V}^{(l)^k} \in \mathbb{R}^{2K_m \times d_V^{(l)}}. \tag{19}$$

The $K_s$ results of the self-attention model $\{\mathbf{Z}^{(l)^1}, \ldots, \mathbf{Z}^{(l)^{K_s}}\}$ are then concatenated as a result vector $\mathbf{z}^{(l)} \in \mathbb{R}^{2K_m d_V^{(l)} K_s}$, which is then input to a fully connected neural network to obtain the neighbor-based embedding of the given entity $e$ in the $l$-th layer, which is defined as follows:

$$\mathbf{e}^{(l)} = \delta(\mathbf{W}_f^{(l)} \mathbf{z}^{(l)} + \mathbf{b}_f^{(l)}) \in \mathbb{R}^{d^{(l)}}, \tag{20}$$

where $\mathbf{W}_f^{(l)} \in \mathbb{R}^{d^{(l)} \times 2K_m d_V^{(l)} K_s}$ and $\mathbf{b}_f^{(l)} \in \mathbb{R}^{d^{(l)}}$ are the weight matrix and bias of the fully connected neural network in the $l$-th layer, respectively.

### 3.3.5 Training Target

After two-layer neighbor aggregation, we finally obtain the neighbor-based embeddings

$\mathbf{e}_{\text{nei}} = \mathbf{e}^{(2)}$ of each entity $e$.

For each triple $(h, r, t)$ in the KG, we use the energy function of TransE as the training target to update the parameters in the neighbor aggregation part, which is defined as follows:

$$E_{\text{NA}}(h, r, t) = |\mathbf{h}_{\text{nei}} + \mathbf{r}_{\text{nei}} - \mathbf{t}_{\text{nei}}|, \tag{21}$$

where $\mathbf{h}_{\text{nei}}$ and $\mathbf{t}_{\text{nei}}$ are the neighbor-based embeddings of $h$ and $t$, respectively; $\mathbf{r}_{\text{nei}} = \mathbf{r}\mathbf{W}_r$ is the transformed embedding of $r$; $\mathbf{r}$ is the base embedding of $r$.

### 3.4 Representation Learning Part

The representation learning part takes the neighbor-based embeddings of entities obtained in the neighbor aggregation part as initial values, and trains a TRL model ConvKB [18] to realize link prediction.

ConvKB is a TRL model based on multi-kernel convolution. Given a triple $(h, r, t)$, the energy function of ConvKB is defined as follows:

$$E_{\text{RL}}(h, r, t) = \left( \Big\|_{m=1}^{\Omega} \text{ReLU}([\mathbf{h}, \mathbf{r}, \mathbf{t}] * \omega^m) \right) \mathbf{W}_{\text{RL}}, \tag{22}$$

where $\omega^m$ is the $m$-th kernel; $\Omega$ is the number of kernels; $\|$ is the concatenation of the vectors; $\mathbf{W}_{\text{RL}}$ is a linear transformation matrix, which transforms the result of the multi-kernel convolution into an energy function value.

### 3.5 Training

In the neighbor aggregation part, the parameters are updated by minimizing a margin-based loss function [4], which is defined as follows:

$$L_{\text{NA}} = \sum_{(h,r,t) \in T} \sum_{(h',r,t') \in T^-} [\gamma + E_{\text{NA}}(h, r, t) - E_{\text{NA}}(h', r, t')]^+, \tag{23}$$

where $[x]^+ = \max(0, x)$ is the maximum between 0 and $x$; $\gamma$ is the margin; $T$ is the set of all triples in the KG; $T^-$ is the set of negative samples, which is defined as follows:

$$\begin{aligned} T^- = &\{(h', r, t) \notin \text{KG} | (h, r, t) \in \text{KG}, h' \in \mathcal{E}\} \\ &\cup \{(h, r, t') \notin \text{KG} | (h, r, t) \in \text{KG}, t' \in \mathcal{E}\}. \end{aligned} \tag{24}$$

In the representation learning part, the parameters are updated by minimizing a regularized negative log-likelihood of the logistic model [18], which is defined as follows:

$$L_{RL} = \sum_{(h,r,t)\in T\cup T^-} \log\left(1 + \exp(E_{\text{RL}}(h,r,t)y(h,r,t))\right) + \frac{\lambda}{2}\|\mathbf{W}_{\text{RL}}\|^2, \qquad (25)$$

where $\lambda$ is the regularization parameter; $y(h,r,t)$ is the label of the triple $(h,r,t)$, which is defined as follows:

$$y(h,r,t) = \begin{cases} 1, (h,r,t) \in T \\ -1, (h,r,t) \in T^- \end{cases}. \qquad (26)$$

The working process of RMNA is shown in Algorithm 1. Firstly, we initialize the base embeddings of relations and entities by TransE, obtain and filter horn rules, and transform valuable multi-hop neighbors into one-hop neighbors through selected horn rules. Then, we use the energy function of TransE as the objective to train the hierarchical neighbor aggregation model to obtain neighbor-based embeddings, which are finally employed to train ConvKB to realize link prediction.

---

**Algorithm 1**: **Model Working Process**

---

**Input**: training set S

**Output**: $\mathbf{h}_{\text{nei}}, \mathbf{r}_{\text{nei}}, \mathbf{t}_{\text{nei}}$

---

1: initialize the base embeddings of relations and entities by TransE

2: use the AMIE algorithm to obtain horn rules $B \to r(e,e')$

3: filter out rules in the form of $r_1(e,e_1) \wedge r_2(e_1,e_2) \wedge \ldots \wedge r_n(e_{n-1},e') \to r(e,e')$

4: match each entity in S with each selected horn rule, and introduce one-hop neighbors

5: **do**

6:   **for** each entity $e$ in $\mathcal{E}$:

7:     dividing one-hop neighbors into original one-hop neighbors and transformed one-hop neighbors

8:     **for** each original one-hop neighbor $(r_o, t_o)$:

9:       calculate hidden vectors by Formula (14)

10:     **for** each introduced one-hop neighbor $(r_t, t_t)$:

11:       calculate hidden vectors by Formula (15)

12:     combine all hidden vectors, forming an input matrix $\mathbf{X}^{(1)}$

13:     calculate query matrix, key matrix, and value matrix by Formulas (16), (17), and (18)

14:     calculate the output of self-attention by Formula (19)

15:     calculate the result of first layer by Formula (20)

16:     change $\mathbf{e}, \mathbf{r}_o, \mathbf{r}_t, \mathbf{t}_o, \mathbf{t}_t$ into $\mathbf{e}^{(1)}, \mathbf{r}_o^{(1)}, \mathbf{r}_t^{(1)}, \mathbf{t}_o^{(1)}, \mathbf{t}_t^{(1)}$ in the second layer

17:     repeat lines 7-15 to get the neighbor-based embedding of $e$: $\mathbf{e}_{nei} = \mathbf{e}^{(2)}$

18:   update parameters by Formula (23)

19: **until** reaching max iterations of the neighbor aggregation part

20: **do**

21:   use neighbor-based embeddings as inputs to train ConvKB

21:   update parameters by Formula (25)

22: **until** reaching max iterations of the representation learning part

## 4  EXPERIMENTS

### 4.1  Datasets

Table 1: The statistics of datasets.

| Dataset | #Rel | #Ent | #Train | #Valid | #Test |
|---|---|---|---|---|---|
| FB15K-237 | 237 | 14541 | 272115 | 17535 | 20466 |
| WN18RR | 11 | 40943 | 86835 | 3034 | 3134 |

FB15K [4] is a subset of Freebase [3], which consists of selected 14951 entities in Freebase and the relations between these entities. FB15K-237 removes redundant relations on the basis of FB15K. WN18RR removes the inverse relations in WN18 [4], which is the subset of WordNet [16]. Table 1 shows the statistics of datasets.

### 4.2  Experimental Settings

We set $l_{max} = 3$ to control the time-consuming and quantity of neighbors. We search head coverage threshold $hc_{min}$, confidence threshold $conf_{min}$ so that the number of two types of neighbors is basically equal. We search dimension of base embeddings $d$ among {50, 100, 200}, learning rate $\lambda$ among {0.0005, 0.001, 0.01}, the number of heads in multi-head attention $K_m$ among {1, 2, 4}, the number of heads in self-attention $K_s$ among {1, 2, 4}, margin $\gamma$ among {1, 2, 3}, dropout probability $P_{\text{dropout}}$ among {0.1, 0.2, 0.3}. The dimensions of hidden vectors in two layers $d^{(1)}$ and $d^{(2)}$, the dimensions of query matrices in two layers $d_Q^{(1)}$ and $d_Q^{(2)}$, the dimensions of key matrices in two layers $d_K^{(1)}$ and $d_K^{(2)}$, the dimensions of value matrices in two layers $d_V^{(1)}$ and $d_V^{(2)}$ are determined by the above parameters.

In the following, we give the parameter settings of RMNA which show the best performance on FB15K-237 and WN18RR:

**FB15K-237** In the rule mining part, we set $l_{\max} = 3$, $hc_{\min} = 0.7$, $conf_{\min} = 0.7$, and $d = 100$. In the neighbor aggregation part, we set $\lambda = 0.001$, $K_m = 2$, $K_s = 4$, $d^{(1)} = 100$, $d^{(2)} = 200$, $d_Q^{(1)} = 25$, $d_Q^{(2)} = 50$, $d_K^{(1)} = 25$, $d_K^{(2)} = 50$, $d_V^{(1)} = 25$, $d_V^{(2)} = 50$, $\gamma = 1$, $P_{\text{dropout}} = 0.3$, and the number of iterations = 2000. In the representation learning part, we set $\lambda = 0.001$, the size of the kernel $1 \times 3$, $P_{\text{dropout}} = 0.3$, and the number of iterations = 150.

**WN18RR** In the rule mining part, we set $l_{\max} = 3$, $hc_{\min} = 0$, $conf_{\min} = 0.2$, and $d = 50$. In the neighbor aggregation part, we set $\lambda = 0.001$, $K_m = 2$, $K_s = 4$, $d^{(1)} = 100$, $d^{(2)} = 200$, $d_Q^{(1)} = 25$, $d_Q^{(2)} = 50$, $d_K^{(1)} = 25$, $d_K^{(2)} = 50$, $d_V^{(1)} = 25$, $d_V^{(2)} = 50$, $\gamma = 1$, $P_{\text{dropout}} = 0.3$, and the number of iterations = 3600. In the representation learning part, we set $\lambda = 0.001$, the size of the kernel $1 \times 3$, $P_{\text{dropout}} = 0.3$, and the number of iterations = 200.

We use Adam optimizer [13] to update parameters. For baselines, we use the best parameters given in their papers.

The code of RMNA is released on GitHub[1]. All the experiments are conducted on a Linux PC with an Intel Core i9-9900K (8 cores, 3.60G HZ) and NVIDIA RTX 2080Ti.

## 4.3 Evaluation Protocol

We use the following common evaluation method [4] to measure the performance of a model on link prediction task: For each test triple in the test set, we replace the head entity (or the tail entity) of the test triple with all other entities in the KG, forming corrupt triples. We rank the test triple and corrupt triples with the energy function in ascending order, and record the rank of the test triple.

---

[1] https://github.com/scd158/RMNA

We choose NARL models A2N, LAN, R-GCN, and KBGAT, and the state-of-the-art TRL models DistMult, ComplEX, and ConvE as baselines. We use Hits@1, Hits@3, Hits@10, and MRR metrics. Hits@n represents the percentage that test triples rank in top n in the ranking; MRR represents the reciprocal of the average ranks of all test triples in the rankings.

We also evaluate four variant models, each of which removes a part of inputs of transformed one-hop neighbors, to justify the effectiveness of the components of RMNA: RMNA_nh, RMNA_nc, RMNA_nl, and RMNA_ns remove the head coverage $hc$, the confidence $conf$, the normalized length $l_{norm}$, and the score $s$ in Formula (5) and Formula (8), respectively.

### 4.4 Results

Table 2 and Table 3 show the performances of RMNA and baselines on FB15K-237 and WN18RR, respectively, from which we can find out:

(1) The performance of LAN is poor on both FB15K-237 and WN18RR, which might be because LAN uses the co-occurrences of the query relation and the neighbor relation as a part of the attention mechanism; however, frequent co-occurrences of two relations do not necessarily mean that they are related. For example, "gender" and "nationality" are two relations that co-occur frequently, as almost every entity whose type is "human" contains these two relations, while they are not related. Wang et al. [27] uses datasets that contain redundant relations to evaluate the performance of LAN. In this case, most of the two relations that frequently co-occur are redundant with each other, and it can be considered that "frequent co-occurrences of two relations mean that they are related" is established. FB15K-237 and WN18RR filter out highly redundant relations, therefore, LAN shows a poor performance.

(2) RMNA has a better performance in most metrics than DistMult, ComlEx, and ConvE on both FB15K-237 and WN18RR, which might be because they are composed of operations that do not support parameter sharing between entities. The embeddings of entities in RMNA share parameters and the connections between entities are strong, which can address the lack of information interaction between entities.

(3) RMNA has a better performance in most metrics than A2N and LAN on both FB15K-237 and WN18RR, which might be because A2N and LAN ignore the information in multi-hop neighbors, while RMNA utilizes the information in multi-hop neighbors.

(4) R-GCN has a worse performance in all metrics than A2N on both FB15K-237 and WN18RR. Although R-GCN classifies the neighbors of an entity according to the type and direction of the neighbor relations, the neighbors in the same classification have the same attention in R-GCN, while A2N measures the weight of each neighbor by a bi-linear attention. In addition, A2N learns a different neighbor-based embedding for each entity under each different relation, while R-GCN learns the same neighbor-based embedding for each entity under different relations.

(5) RMNA has a better performance in most metrics than KBGAT on both FB15K-237 and WN18RR, which might be because KBGAT splits a complete multi-hop neighbor into several one-hop neighbors, destroying the completeness of multi-hop neighbors. On the contrary, RMNA transforms valuable multi-hop neighbors into one-hop neighbors that are semantically similar to the corresponding multi-hop neighbors, so that the completeness of multi-hop neighbors can be ensured.

Table 2: The performances of RMNA and baselines on FB15K-237.

| Metric(%) | MRR | Hits@1 | Hits@3 | Hits@10 |
|---|---|---|---|---|
| A2N | 31.4 | 22.9 | 35.0 | 49.1 |
| LAN | 21.1 | 8.5 | 16.8 | 31.2 |
| R-GCN | 24.8 | 15.7 | 26.3 | 42.4 |
| KBGAT | 44.6 | 36.2 | 48.3 | 60.8 |
| DistMult | 29.0 | 20.8 | 31.8 | 45.4 |
| ComplEx | 28.2 | 20.0 | 30.9 | 45.0 |
| ConvE | 30.9 | 22.3 | 33.7 | 48.7 |
| RMNA | 45.9 | 38.0 | 49.5 | 61.6 |

Table 3: The performances of RMNA and baselines on WN18RR.

| Metric(%) | MRR | Hits@1 | Hits@3 | Hits@10 |
|---|---|---|---|---|
| A2N | 44.2 | 41.3 | 46.8 | 52.3 |
| LAN | 33.6 | 18.4 | 25.7 | 40.1 |
| R-GCN | 11.5 | 9.1 | 13.6 | 21.4 |
| KBGAT | 43.9 | 36.3 | 48.0 | 58.1 |
| DistMult | 43.0 | 40.3 | 46.5 | 51.6 |
| ComplEx | 43.7 | 41,1 | 47.2 | 52.8 |
| ConvE | 44.7 | 40.3 | 46.0 | 53.5 |
| RMNA | 44.1 | 36.0 | 48.8 | 58.4 |

Table 4 and Table 5 show the performances of RMNA and its variant models on FB15K-237 and WN18RR, respectively, from which we can find out:

(1) RMNA has a better performance in most metrics than RMNA_nh on both FB15K-237 and WN18RR. Head coverage represents the probability that when a transformed one-hop neighbor appears, the corresponding valuable multi-hop neighbor also appears in the KG, which can measure the reliability of introducing a transformed one-hop neighbor. The results show the effectiveness of measuring the weights of transformed one-hop neighbors by head coverage.

(2) RMNA has a better performance in all metrics than RMNA_nc on both FB15K-237 and WN18RR. Confidence represents the probability that a transformed one-hop neighbor also appears, when the corresponding valuable multi-hop neighbor appears in the KG, which can measure the reliability of introducing a transformed one-hop neighbor. The results show the effectiveness of measuring the weights of transformed one-hop neighbors by confidence.

(3) RMNA has a better performance in most metrics than RMNA_nl on both FB15K-237 and WN18RR. Normalized length represents the normalized length of the corresponding horn rule, i.e., the normalized length of a valuable multi-hop neighbor. The results show that RMNA can learn the relationship between the length of the valuable multi-hop neighbors and the weights of transformed one-hop neighbors.

(4) RMNA has a better performance in all metrics than RMNA_ns on both FB15K-237 and WN18RR. The score is calculated through the energy function of TransE on the original KG, which can measure the fit of a transformed one-hop neighbor to the original KG. The results show the effectiveness of measuring the weights of transformed one-hop neighbors by score.

Table 4: The performances of RMNA and its variant models on FB15K-237.

| Metric(%) | MRR  | Hits@1 | Hits@3 | Hits@10 |
|-----------|------|--------|--------|---------|
| RMNA_nh   | 44.1 | 34.4   | 49.2   | 61.4    |
| RMNA_nc   | 44.3 | 36.7   | 47.5   | 59.1    |
| RMNA_nl   | 44.3 | 37.3   | 46.9   | 58.0    |
| RMNA_ns   | 43.5 | 33.6   | 48.9   | 61.2    |
| RMNA      | 45.9 | 38.0   | 49.5   | 61.6    |

Table 5: The performances of RMNA and its variant models on WN18RR.

| Metric(%) | MRR | Hits@1 | Hits@3 | Hits@10 |
|---|---|---|---|---|
| RMNA_nh | 44.0 | 35.9 | 48.8 | 58.2 |
| RMNA_nc | 42.3 | 34.1 | 47.4 | 56.9 |
| RMNA_nl | 44.1 | 35.7 | 48.5 | 58.3 |
| RMNA_ns | 43.9 | 35.8 | 48.6 | 58.3 |
| RMNA | 44.1 | 36.0 | 48.8 | 58.4 |

## 5 CONCLUSIONS AND FUTURE WORK

In this paper, we propose a NARL model RMNA, which transforms valuable multi-hop neighbors into one-hop neighbors with similar semantics by selected horn rules, separately and hierarchically aggregates the original one-hop neighbors and the transformed one-hop neighbors of an entity by the multi-head attention mechanism, and aggregates the heads by the self-attention mechanism to obtain the neighbor-based embedding of the entity. In experiments, RMNA shows competitive performance, justifying that RMNA can utilize information in neighbors effectively.

In the future, we will continue our study in the following aspects: (1) The hierarchical neighbor aggregation model only consists of two layers, we will increase the number of layers to further exploit the information in neighbors. (2) As the length of horn rules increases, the overhead of rule mining increases exponentially. We will jointly learn embeddings and search horn rules, and prune through the learned embeddings to reduce the search space.

## 6 ACKNOWLEDGEMENT

This work was funded by the National Key Research and Development Program of China (No. 2018YFB0505000) and the Fundamental Research Funds for the Central Universities (No. 2020QNA5017).